\documentclass{article}

\usepackage[final]{neurips_2025}

\usepackage[utf8]{inputenc} % allow utf-8 input
\usepackage[T1]{fontenc}    % use 8-bit T1 fonts
\usepackage{hyperref}       % hyperlinks
\usepackage{url}            % simple URL typesetting
\usepackage{booktabs}       % professional-quality tables
\usepackage{amsfonts}       % blackboard math symbols
\usepackage{nicefrac}       % compact symbols for 1/2, etc.
\usepackage{microtype}      % microtypography
\usepackage{xcolor}         % colors

\usepackage{microtype}
\usepackage{longtable}     % 支持跨页表格
\usepackage{tabularx}      % 智能列宽调整
\usepackage{ragged2e}      % 更好的对齐控制
\usepackage{booktabs}      % 专业表格线
\usepackage{ltablex}       % 结合longtable和tabularx
\usepackage{geometry}      % 调整页边距
\usepackage{graphicx} 
\usepackage{amssymb}         % 提供数学符号（如 ⇒）
\usepackage{amsmath}         % 数学扩展（有些箭头需要）
\usepackage{caption}
\usepackage{adjustbox}
\usepackage{threeparttable}
\usepackage{afterpage}
\usepackage{siunitx} % 加入导言区

\usepackage{booktabs}
\usepackage{tabularx}
\usepackage{caption}
\usepackage{array}
\usepackage{xspace}
\usepackage{CJKutf8}

\title{AutoEvoEval: An Automated Framework for Evolving Close-Ended LLM Evaluation Data}

\author{
  Jiaru Wu \\
  Sun Yat-sen University \\
  Zhuhai, China\\
  \texttt{wujr33@mail2.sysu.edu.cn}
  \And
  Mingwei Liu\thanks{*} \\
  Sun Yat-sen University \\
  Zhuhai, China\\
  \texttt{liumw26@mail.sysu.edu.cn}
}

\newcommand{\app}{AutoEvoEval\xspace}
\newcommand{\parabf}[1]
{\paragraph{\textbf{#1}}}

\begin{document}

\maketitle

\begin{abstract}
Large language models (LLMs) have shown remarkable performance on various tasks, but existing evaluation benchmarks are often static and insufficient to fully assess their robustness and generalization in realistic scenarios. Prior work using evolutionary or adversarial data augmentation has improved evaluation diversity but lacks systematic control over perturbation types and multi-step complexity, limiting comprehensive robustness analysis. To address these gaps, we propose AutoEvoEval, an evolution-based evaluation framework for close-ended tasks such as multi-choice question answering. AutoEvoEval introduces 22 interpretable atomic evolution operations and supports multi-round compositions, enabling controlled generation of diverse, challenging, and realistic test samples.
We conduct extensive experiments addressing four research questions on a broad set of open- and closed-source LLMs. Our results show that atomic operations cause an average accuracy drop of 7.283\%, with structure-disrupting or misleading semantic edits causing the largest declines. Model sensitivities vary significantly for the same perturbation, and combining multiple evolution steps amplifies adversarial effects by up to 52.932\%. These findings suggest current benchmarks may overestimate true model generalization and emphasize the need for evolution-aware robustness evaluation. Code and resources are available at: \url{https://github.com/SYSUSELab/AutoEvoEval}.

\end{abstract}

\section{Introduction}
With the rapid advancement of large language models (LLMs), their impressive performance on various natural language processing tasks has drawn widespread attention. Evaluating these models is crucial not only to understand their strengths and limitations but also to guide future model improvement. However, current evaluation paradigms often rely on static benchmarks that fail to capture the diversity and complexity of real-world inputs, limiting their effectiveness and potentially overestimating model generalization.

Recent work has begun to address these limitations by introducing evolutionary or dynamic datasets that automatically generate and modify test samples to increase challenge and diversity~\cite{Li2023BeyondSD}. For example, Alcuna et al.~\cite{Yin2023ALCUNALL} generates adversarial questions by applying perturbations at multiple linguistic levels, such as character, word, and sentence, to assess robustness. Despite these advances, existing methods face critical challenges: (1) they lack a comprehensive and systematic taxonomy of evolution operations to fully probe model reasoning and generalization; (2) they offer limited control over the types and complexity of data evolution; and (3) evaluation often remains task- or dataset-specific, reducing broader applicability.

To address these gaps, we propose \app, a systematic evolution-based evaluation framework tailored for close-ended tasks such as multi-choice question answering. \app defines 22 interpretable atomic evolution operations that simulate realistic semantic, syntactic, and structural perturbations. These operations can be composed into multi-round evolution chains, allowing controlled generation of progressively more challenging test samples. This enables thorough assessment of LLM robustness, adaptability, semantic understanding, and sensitivity to noise or adversarial manipulations.

We conduct extensive experiments addressing four research questions (RQ) to evaluate how different types and lengths of evolutionary transformations affect a diverse set of LLMs. Our results show that atomic operations disrupting logical structure or injecting misleading semantics cause the largest accuracy drops, with an average decline of approximately 30.748\% across models, making them effective robustness stress tests. Model sensitivity to the same perturbations varies widely, with performance differences up to 36.49 points between models, underscoring the need for fine-grained evaluation. Moreover, combining multiple evolutionary steps amplifies adversarial effects, resulting in accuracy degradation up to 50.036\% worse than single-step perturbations, revealing the compounded vulnerability of LLMs. Finally, current benchmarks may overestimate true generalization capabilities, as even minor atomic edits lead to performance drops averaging 7.283\%.

Our main contributions are summarized as follows:

We introduce \app, a comprehensive evolution-based evaluation framework featuring a rich taxonomy of 22 atomic operations and multi-round composition, enabling more realistic and challenging robustness assessment.
We systematically investigate the impact of diverse evolution types and chain lengths on multiple LLMs, providing novel insights into model vulnerabilities and evaluation best practices.
We empirically demonstrate that structural and semantic disruptions are particularly effective in revealing robustness weaknesses, and that long evolution chains exacerbate performance degradation.

\section{Related Work}

\textbf{Close-Ended Benchmarks for Knowledge Evaluation.}
With the rapid advancement of LLMs, increasing attention has been devoted to systematically and comprehensively evaluating their capabilities. A core aspect of this evaluation is the measurement of a model’s knowledge capacity, which is commonly conducted through close-ended multiple-choice benchmarks due to their structured format, ease of automatic grading, and reproducibility.

These benchmarks can be broadly categorized into two types. General-purpose benchmarks, such as \textbf{MMLU}\citep{hendrycks2021measuringmassivemultitasklanguage}, \textbf{C-Eval}\citep{huang2023cevalmultilevelmultidisciplinechinese}, and \textbf{ARC}~\citep{clark2018thinksolvedquestionanswering}, are designed to assess a model's broad knowledge coverage across a wide range of subjects and reasoning types. MMLU, for example, covers 57 tasks spanning humanities, STEM, and social sciences, and is widely adopted as a standard for evaluating general knowledge and reasoning in LLMs. C-Eval extends this paradigm to Chinese-language contexts, covering 52 disciplines aligned with university curricula.

In contrast, domain-specific benchmarks focus on professional knowledge in specialized fields. Examples include \textbf{MedMCQA}\citep{pal2022medmcqalargescalemultisubject}, which targets medical entrance exam questions in India, and \textbf{ScienceQA}\citep{lu2022learnexplainmultimodalreasoning}, which evaluates science understanding with multimodal questions incorporating text and diagrams. These benchmarks require more precise domain knowledge and are particularly useful for assessing the LLM’s ability to operate in high-stakes or expert-level scenarios.

While these benchmarks have significantly contributed to the standardization of LLM evaluation, they are predominantly constructed in a static manner. As LLMs continue to evolve and gain exposure to public datasets, concerns have been raised regarding potential data contamination, memorization, and diminishing benchmark utility. These limitations call for more dynamic and adaptive evaluation approaches, especially in the context of assessing deep and nuanced knowledge capabilities.

\textbf{Evolving Evaluation Data for LLMs.}
To address the inherent limitations of static, close-ended benchmarks, recent studies have explored automatic data generation and evolutionary techniques aimed at increasing the diversity, complexity, and adaptability of evaluation datasets.

For instance, \textbf{Alcuna} introduces the concept of \textit{knowledge genes}, leveraging knowledge graphs to construct adversarial yet semantically valid questions that challenge the factual reasoning capabilities of LLMs~\citep{yin2023alcunalargelanguagemodels}. Other perturbation-based approaches operate at different granularity levels: character-level noise injection~\citep{8424632}, word-level adversarial replacements~\citep{li-etal-2023-adversarial}, and sentence-level rewrites targeting statistical biases in model predictions~\citep{lin2021usingadversarialattacksreveal}. In addition, \textbf{Perteval}~\citep{li2024pertevalunveilingrealknowledge} offers a toolkit for generating semantically equivalent variants via controlled input reformatting, supporting contamination-resilient evaluation.

Although these methods help reveal model weaknesses and improve robustness testing, they are often limited to single-step or isolated perturbations, lacking systematic and compositional capabilities. To overcome these limitations, we propose an evolutionary evaluation and governance framework that composes atomic evolution strategies into multi-step chains. This enables the generation of more diverse, realistic, and challenging test cases, thereby strengthening the evaluation of LLMs' knowledge capacity under dynamic and adversarial conditions.

\section{Methodology}

\begin{figure}[htpb]
    \centering
    \includegraphics[width=1.0\linewidth]{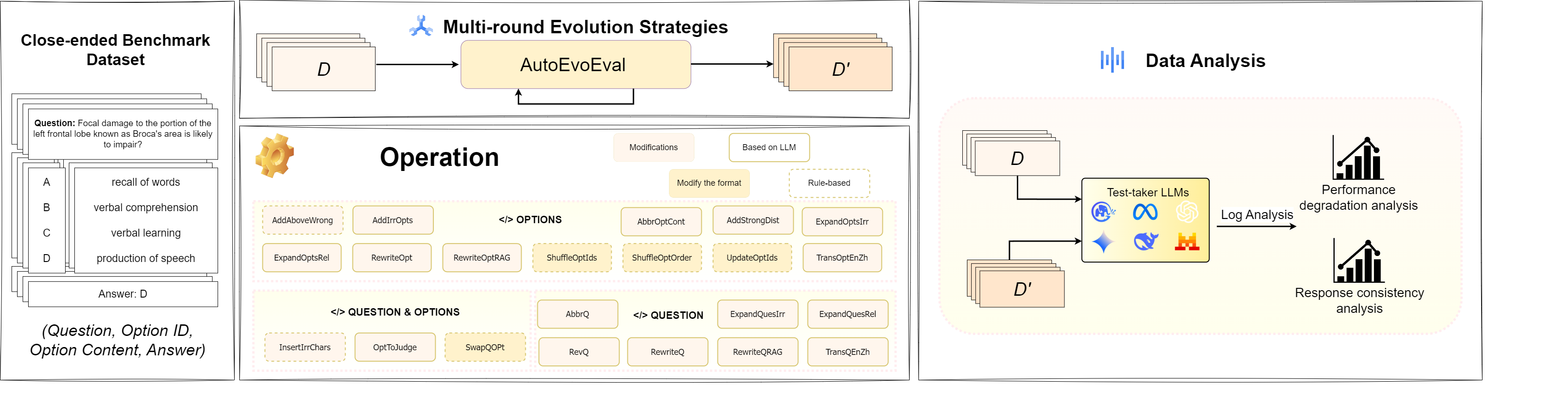}
    \caption{Overview of the AutoEvoEval Framework: AutoEvoEval generates a perturbed dataset D' from a closed benchmark dataset D using atomic evolution operations and multi-round iterations. It then evaluates model capability and robustness by analyzing performance degradation and response consistency.}
    \label{fig:methods}
\end{figure}

Given a close-ended evaluation dataset---where each instance is defined as a tuple $(q, \{(id_i, o_i)\}_{i=1}^n, a)$, consisting of a \textit{question} $q$, a set of answer \textit{options} $\{(id_i, o_i)\}$ (each with an identifier $id_i$ and content $o_i$), and the \textit{correct answer} $a$---we introduce \textbf{\app}, a unified and extensible framework (illustrated in Fig.~\ref{fig:methods}), designed to automatically evolve such datasets for more diverse, challenging, and robust evaluation of LLMs.

Our framework supports both \textit{single-step} and \textit{multi-round} evolution strategies, integrating a collection of atomic evolution operations that target the semantics and/or formatting of the input. These atomic operations can be flexibly composed into evolution chains, allowing different perturbation strategies to be applied iteratively. Our design maintains full compatibility with standard close-ended formats (e.g., multiple-choice, true/false, single-label judgment) and enables fine-grained control over the type and difficulty of the evolved instances.

Below, we provide detailed descriptions of the atomic evolution operations, the multi-round iterative evolution process, and the evaluation strategies.

\subsection{Atomic Evolution Operations}

\begin{table}[ht]
\caption{Complete Evolutionary Methods for Generating Question Variants}
\label{tab:evolution_methods}
\scriptsize
\setlength{\tabcolsep}{3.5pt}
\renewcommand{\arraystretch}{0.92}
\begin{tabularx}{\textwidth}{@{}>{\raggedright}p{1.8cm}X cc>{\ttfamily}l>{\ttfamily}l@{}}
\toprule
\multicolumn{1}{c}{\textbf{Methods}} & \textbf{Definition} & \textbf{Type1} & \textbf{Type2} & \textbf{Original} & \textbf{Evolved} \\ 
\midrule
Origin & Original question from dataset & -- & -- & Q:xxx (A,B,C,D)xxx & -- \\
AbbrOptCont & Shortens options using concise language & LLM & Opt & (A,B,C,D)xxx & (A,B,C,D)xx \\
AbbrQ & Condense the question and express it concisely & LLM & Q & Q:xxx & Q:xx \\
AddAboveWrong & Replace correct option with "None of the above" & Rule & Opt & (A,\_B,C,D)xxx & (A,C,D)xxx, (\_B)None... \\
AddIrrOpts & Add unrelated options & LLM & Opt & (A,B,C,D)xxx & (A,B,C,D)xxx,(E)\textit{IR}xxx \\
AddStrongDist & Add plausible distractors & LLM & Opt & (A,B,C,D)xxx & (A,B,C,D)xxx,(E)\textit{R}xxx \\
OptToJudge & Combine correct and incorrect options to form true/false questions & LLM & Both & Q:xxx (A,B,\_C,D)xxx & J\_A + J\_C \\
ExpendOptsIrr & Inject unrelated info into options & LLM & Opt & (A,B,C,D)xxx & (A,B,C,D)\textit{IR}xxxxx \\
ExpandOptsRel & Add supporting info to options & LLM & Opt & (A,B,C,D)xxx & (A,B,C,D)\textit{R}xxxxx \\
ExpandQuesIrr & Add irrelevant content to question & LLM & Q & Q:xxx & Q:\textit{IR}xxxxx \\
ExpandQuesRel & Append relevant knowledge to question & LLM & Q & Q:xxx & Q:\textit{R}xxxxx \\
InsertIrrChars & Insert random non-disruptive characters & Rule & Both & Q:xxx (A,B,C,D)xxx & \texttt{Q:x\char36{}x\char36{}x (A,B,C,D)x\char36{}x\char36{}x} \\
RevQ & Reverse question logic & LLM & Q & Q:xxx (A,B,C,\textbf{D})xxx & Q:xxx' (\textbf{A},\textbf{B},\textbf{C},D)xxx \\
RewriteOpt & Semantic-preserving option rewriting & LLM & Opt & (A,B,C,D)xxx & (A,B,C,D)xxx' \\
RewriteOptRAG & RAG-based option rewriting & LLM & Opt & (A,B,C,D)xxx & (A,B,C,D)xxx' + K \\
RewriteQ & Semantic-preserving question rewriting & LLM & Q & Q:xxx & Q:xxx' \\
RewriteQRAG & RAG-enhanced question rewriting & LLM & Q & Q:xxx & Q:xxx' + K \\
ShuffleOptIds & Randomly reassigns option identifiers & Rule & Opt & (A,B,C,D) & (C,A,D,B) \\
ShuffleOptOrder & Randomize option positions & Rule & Opt & (A)xx,(B)xxx... & (B)xxx,(A)xx... \\
SwapQOpt & Swap question-option positions & Rule & Both & Q:xxx (A,B,C,D)xxx & (A,B,C,D)xxx Q:xxx \\
TransOptEnZh & Translate answer options between English and Chinese & LLM & Opt & (A,B,C,D)xxx & (A,B,C,D)xxx' \\
TransQEnZh & Translate the question between English and Chinese & LLM & Q & Q:xxx & Q:xxx' \\
UpdateOptIds & Assign a random value to the option id & Rule & Opt & (A,B,C,D) & (E,U,M,V) \\
\bottomrule
\end{tabularx}

\vspace{2mm}
\parbox{\textwidth}{\scriptsize
\textit{Legend:} 
\textit{IR}: Irrelevant information, 
\textit{R}: Related information, 
K: RAG knowledge, 
J: Judgment format,
\texttt{\$}: randomly inserted non-disruptive characters,
\texttt{xxx/xx/xxxxx}: variable-length text segments,
\texttt{()}: Options,
\texttt{Q}: Question,
\_Id: Correct Answer,
': Some changes
}
\end{table}

We define each close-ended evaluation instance as a tuple $(q, \{(id_i, o_i)\}_{i=1}^n, a)$, where $q$ denotes the question, $\{(id_i, o_i)\}$ is a set of option identifiers and their corresponding contents, and $a$ is the correct answer.

To enhance evaluation diversity and robustness, \textbf{\app} employs a set of \textit{atomic evolution operations} at three granularity levels: question, option, and joint question-option. Table~\ref{tab:evolution_methods} summarizes these operations with examples. Each is designed to preserve the original question’s solvability while modifying format, semantics, or distractors to challenge LLMs beyond memorization. Detailed descriptions are in Appendix~\ref{appendix:atomic_operations} and our codebase. Operations are composable, reversible, and can be applied independently or sequentially. We ensure transformed instances remain answerable with updated labels as needed, supporting both controlled perturbations and large-scale automated evolution.

\parabf{Question-Level Evolutions.}  
At the question level, we focus on evolving the question text itself while preserving its core knowledge and answerability. This includes operations such as rewriting, expanding, translating or compressing the question content to simulate diverse linguistic expressions and formats. These evolutions are primarily realized through LLMs guided by carefully designed prompts.

We implement two main approaches: one fully leverages the generative capabilities of large models to perform question evolution independently, and the other employs a Retriever-Augmented Generation (RAG) framework. The RAG-based method first retrieves relevant external knowledge from a domain-specific knowledge base—such as a collection of historical texts for history-related datasets—and then uses this context to guide the question rewriting. This enables incorporating richer, context-aware variations beyond the model’s inherent knowledge.

\parabf{Option-Level Evolutions.}  
Option-level evolutions focus on modifying the answer choices’ content and format to increase diversity and evaluation challenge. These include shortening or rewriting option texts, translating them, introducing various types of distractors, and altering the representation of the correct answer, such as converting it to ``None of the above.'' Similar to question-level evolutions, we utilize large language models with prompt-guided generation, alongside RAG-based rewriting that integrates external knowledge to ensure context relevance.

Compared to question-level changes—which primarily alter the question stem's linguistic form without changing its core knowledge—option-level evolutions emphasize challenging the model’s discrimination ability among plausible or confusing choices. This level includes both semantic modifications (e.g., adding misleading distractors) and format transformations (e.g., shuffling option orders or updating option labels), aiming to expose models’ biases or superficial pattern learning.

\parabf{Joint Question-Option Level Evolutions.}  
Joint evolutions alter the combined structure or presentation of the question and its options. Examples include swapping the positions of the question stem and options, or inserting irrelevant characters to simulate noisy or user-like input. Another common transformation is converting multiple-choice questions into a series of true/false judgment questions—one per option—where all judgments must be correct for the overall answer to be considered correct. These changes disrupt fixed prompt formats and increase evaluation granularity, testing the model’s adaptability to structural and format variations.

\subsection{Multi-Round Iterative Evolution Process}
Our framework applies atomic evolution operations iteratively over multiple rounds to progressively increase the diversity and complexity of the evaluation dataset. In each round, a subset (or all) of the instances is selected, and one or more atomic operations are applied—either independently or in combination. The output of one round serves as the input for the next, enabling cumulative transformations that simulate real-world variations and present increasing challenges to LLMs.

To ensure quality, each round includes correctness checks to verify that evolved instances remain solvable and that answer labels are updated as needed. The evolution process continues until a predefined diversity threshold is met or the maximum number of iterations is reached. Notably, certain operations—such as converting multiple-choice questions into judgment-format ones—are only applied in the final round to avoid early disruption of structure and semantics.

\subsection{Evaluation Strategies}
The evolved datasets produced by our framework support a range of evaluation strategies for more comprehensive assessment of LLM capabilities. Compared to the original data, the evolved instances are intentionally more challenging, making them suitable for probing the limits of model understanding and generalization.

Specifically, the evolved datasets can be used to:

Assess performance degradation on semantically equivalent but perturbed questions, revealing robustness under surface-level variations.
Measure consistency in answers across different rounds of evolution to evaluate stability under incremental transformations.
Analyze sensitivity to specific evolution operations, quantifying how various types of perturbations (e.g., rewording, distractors, format changes) affect model accuracy.
Compare capability across rounds, observing whether the model's performance declines, remains stable, or improves as the difficulty and variation increase.

These strategies enable more nuanced understanding of LLM behavior beyond single-shot accuracy, offering insights into their true reasoning ability, generalization, and robustness under realistic perturbations.

\section{Experiment}
\label{sec:experiment}
% \textcolor{red}{We first introduce the experimental setup, followed by evaluations of evolution strategies, multi-round effects, and model performance.} %这句话要改一改
\subsection{Setting}
We evaluate on four multiple-choice datasets across distinct domains: \textit{College Mathematics} (C-Math), \textit{World History} (W-History), \textit{Professional Psychology} (P-Psychology), and \textit{Professional Medicine} (P-Medicine) from the Massive Multitask Language Understanding (MMLU)~\cite{hendrycks2021measuringmassivemultitasklanguage} for evaluation. \textbf{DeepSeek-V3}~\cite{deepseekv3} is used to generate evolved data. Evaluation is conducted via API calls to several state-of-the-art LLMs, including both proprietary (e.g., GPT-4, Gemini) and open-source (e.g., GLM-4, LLaMA-3.1) models. Full model list is in Appendix~\ref{appendix:models}. All models are tested on the same original and evolved data using a unified zero-shot prompt. For standard multiple-choice questions, accuracy is computed by comparing the model’s output with the correct option(s). For the \textbf{ConvertOptionsToJudge} type, each question is split into two true/false items—one for the correct option, one for the distractors—and only if both are correctly answered is the original item considered correct. For questions with multiple correct answers, we assign 1.0 for fully correct answers, 0.3 if only some correct options are selected (and no incorrect ones), and 0.0 if any incorrect option is chosen.

\subsection{RQ1: Impact of Atomic Evolutions on Model Performance}
\label{sec:rq1}

\parabf{Design.} We evaluate the impact of each of the 22 atomic evolution operations (Table~\ref{tab:rq1_all}) applied independently on the original datasets. Each operations alters the question, options, or both. Variants are tested on all LLMs, and accuracy drop ($\Delta$Acc) quantifies how much each method challenges model performance.

\begin{table}[htbp]
\centering
\caption{Mean Accuracy Drop per Evolution Operation (Averaged across LLMs)}
\small
\label{tab:rq1_all}
\begin{tabular}{
    l
    S[table-format=2.3]
    S[table-format=2.3]
    S[table-format=2.3]
    S[table-format=2.3]
    S[table-format=2.3]
}
\toprule
\textbf{Method} & \textbf{History} & \textbf{Math} & \textbf{Medicine} & \textbf{Psychology} & \textbf{AVG} \\
\midrule
AbbrOptCont            & -1.788  & -0.125  & -0.983   & -3.085     & -1.495           \\
\textbf{AbbrQ}         & -5.264  & -3.000  & -8.746   & -9.254     & \textbf{-6.566}  \\
\textbf{AddAboveWrong}       & -19.135 & -25.000 & -41.966  & -36.890    & \textbf{-30.748} \\
AddIrrOpts      & -3.950  & -4.750  & -2.195   & -4.175     & -3.768           \\
AddStrongDist               & -7.484  & -4.875  & -2.624   & -3.894     & -4.719           \\
\textbf{OptToJudge}       & -17.663 & -14.750 & -9.431   & -14.863    & \textbf{-14.177} \\
ExpendOptsIrr      & 4.167   & 19.375  & 8.591    & 5.536      & 9.417            \\
ExpandOptsRel        & 5.063   & 17.125  & 8.896    & 6.290      & 9.344            \\
ExpandQuesIrr     & 0.570   & -3.375  & -1.409   & -2.493     & -1.677           \\
ExpandQuesRel       & 3.228   & -3.500  & 2.458    & 4.352      & 1.634            \\
\textbf{InsertIrrChars} & -6.377  & -10.800 & -14.206  & -13.153    & \textbf{-11.134} \\
\textbf{RevQ}            & -49.884 & -26.375 & -49.569  & -49.248    & \textbf{-43.769} \\
RewriteOpt               & -3.107  & -0.375  & -6.159   & -8.233     & -4.468           \\
\textbf{RewriteOptRAG}     & -12.447 & {-}     & -20.548  & -20.753    & \textbf{-17.916} \\
RewriteQ                     & -1.719  & -1.000  & -1.813   & -0.859     & -1.348           \\
RewriteQRAG             & -0.717  & {-}     & -4.073   & 0.361      & -1.476           \\
ShuffleOptIds                    & -3.924  & -2.625  & -2.295   & -4.393     & -3.309           \\
ShuffleOptOrder                 & -2.896  & -5.500  & -2.495   & -4.674     & -3.891           \\
SwapQOpt            & -12.368 & -13.000 & -11.269  & -16.864    & \textbf{-13.375} \\
\textbf{TransOptEnZh}       & 0.016   & -3.125  & -2.718   & -6.177     & -3.001           \\
\textbf{TransQEnZh}      & -4.198  & -7.250  & -10.534  & -10.602    & \textbf{-8.146}  \\
\textbf{UpdateOptIds}            & -4.789  & -3.625  & -4.055   & -10.085    & \textbf{-5.639}  \\
\midrule
\textbf{AVG}                        & \textbf{-6.576} & \textbf{-4.828} & \textbf{-8.052} & \textbf{-9.234} & \textbf{-7.283} \\
\bottomrule
\end{tabular}
\end{table}

\parabf{Impact by Evolution Operation Type.}
Table~\ref{tab:rq1_all} shows that the effectiveness of atomic evolutions varies widely.
\textbf{Operations that alter the logical structure of questions or mislead option interpretation}---such as \texttt{RevQ} (-43.769) and \texttt{AddAboveWrong} (-30.748)—--lead to the most substantial accuracy drops. Similarly, converting options into judgment statements (\texttt{OptToJudge}, -14.177) significantly impairs model reasoning.

Even surface-level modifications, like \texttt{InsertIrrChars} (-11.134), cause noticeable performance degradation, revealing model sensitivity to formatting noise.
In contrast, knowledge injection operations have varying effects—retrieval-based changes (\texttt{RewriteOptRAG}, -17.916) degrade performance more than simple rewriting (\texttt{RewriteOpt}, -4.468), possibly due to the introduction of distracting or conflicting information. Interestingly, some evolutions improve accuracy, such as \texttt{ExpandOptsRel} (+9.344), likely because the added content provides useful contextual cues, even if unintended.

\textbf{Overall, operations that disrupt logical structures or inject misleading semantics are most effective for stress-testing model robustness.} The average accuracy drop across all atomic operations is 7.283, demonstrating their utility in robustness evaluation.

\parabf{Model Sensitivity Differences.}
\textbf{Different models show varied sensitivity to the same atomic operation.} For instance, \texttt{InsertIrrChars} causes a large accuracy drop for \textbf{GLM-4}, but has a much smaller effect on \textbf{DeepSeek-R1}, likely due to differences in tokenization strategies or input preprocessing. Similarly, translation-based evolutions like \texttt{TransOptEnZh} and \texttt{TransQEnZh} impact \textbf{DeepSeek-R1} and \textbf{DeepSeek-V3} more than they do \textbf{GPT-3.5} and \textbf{GPT-4}, possibly because the DeepSeek models have weaker multilingual capabilities or less bilingual training exposure. Detailed per-dataset results are available in Appendix~\ref{rq1_dataset__detail_results}.

\subsection{RQ2: How does the evolution strategy affect the consistency of LLM responses?}

\parabf{Baseline.}
We compare our framework with \textbf{PertEval}~\citep{li2024pertevalunveilingrealknowledge}, which generates semantically equivalent variants through controlled input reformatting for contamination-resilient evaluation of close-ended tasks. PertEval includes a set of simple transformation methods (e.g., option shuffling, format changes) aimed at testing surface-level robustness.

Our framework fully \textbf{implements all PertEval strategies} and significantly extends them with a broader set of \textbf{semantic- and structure-level evolution methods}. Specifically, we reproduce the following PertEval methods within our system: \textit{KnInvPara} as \textit{RewriteQ}, \textit{OptionPerm} as \textit{ShuffleOptOrder}, \textit{OptionForm} as \textit{InsertIrrChars}, \textit{OptionCasar} as \textit{UpdateOptIds}, \textit{ChangeType} as \textit{OptToJudge}, and \textit{SwapPos} as \textit{SwapQOpt}. Beyond these, we introduce additional transformations (marked as ``Others'' in Table~\ref{rq2_ROP_results}) that combine structural edits and semantic-level perturbations, offering more diverse and challenging robustness scenarios.

\parabf{Metric.}
We adopt the \textbf{Recall of Performance (ROP)} from PertEval to quantify LLM robustness under perturbations. ROP measures the proportion of correct answers that remain correct after input evolution:
\[
\text{ROP} = \frac{\text{CC}}{\text{CC} + \text{IC}}
\]
where \textbf{CC} denotes cases consistently correct after perturbation, and \textbf{IC} counts those that turn incorrect. A higher ROP indicates stronger robustness.

\begin{table}[htpb]
\centering
\small % 或 \scriptsize 更小
\caption{
Recall of Performance of LLMs. 
Top row: Perturbation types — KIP: Knowledge-invariant paraphrasing, OP: Option permutation, OF: Option format refactoring, OC: Option ID shifting, CT: Question type changing, SP: Question position swapping. "Others" include RQ, SOO, IIC, UOI, COTJ, SQWO. * from~\cite{li2024pertevalunveilingrealknowledge}
}
\label{rq2_ROP_results}
\begin{tabular*}{\linewidth}{@{\extracolsep{\fill}}l*{7}{S[table-format=1.3]}S[table-format=1.3]}
\toprule
\textbf{PertEval} & \textbf{KIP} & \textbf{OP} & \textbf{OF} & \textbf{OC} & \textbf{CT} & \textbf{SP} & \textbf{Others} & \textbf{AVG} \\
\midrule
GPT-4       & 0.879*  & 0.906*  & 0.960*  & 0.934*  & 0.922*  & 0.799*  & --     & 0.900* \\
\midrule
\textbf{(Ours)}   & \textbf{RQ} & \textbf{SOO} & \textbf{IIC} & \textbf{UOI} & \textbf{COTJ} & \textbf{SQWO} & \textbf{Others} & \textbf{AVG} \\
\midrule
DeepSeek-R1       & 0.912  & 0.930  & 0.928  & 0.926  & 0.846  & 0.908  & 0.857  & 0.871 \\
DeepSeek-V3       & 0.909  & 0.832  & 0.818  & 0.882  & 0.643  & 0.752  & 0.853  & 0.840 \\
Gemini-1.5        & 0.897  & 0.810  & 0.810  & 0.777  & 0.673  & 0.608  & 0.766  & 0.765 \\
GLM-4             & 0.897  & 0.860  & 0.698  & 0.900  & 0.567  & 0.717  & 0.804  & 0.795 \\
Llama-3.1         & 0.579  & 0.236  & 0.251  & 0.161  & 0.410  & 0.201  & 0.460  & 0.418 \\
Mistral-small     & 0.833  & 0.801  & 0.609  & 0.804  & 0.574  & 0.518  & 0.747  & 0.731 \\
GPT-3.5           & 0.896  & 0.872  & 0.699  & 0.914  & 0.598  & 0.721  & 0.804  & 0.799 \\
GPT-4             & 0.909  & 0.855  & 0.733  & 0.885  & 0.586  & 0.734  & 0.805  & 0.799 \\
\midrule
AVG               & 0.854  & 0.774  & 0.693  & 0.781  & 0.612  & 0.645  & --     & --    \\
\bottomrule
\end{tabular*}
\end{table}

\parabf{Results.}
As shown in Table~\ref{rq2_ROP_results}, our framework achieves comparable or better ROP scores for overlapping operations, indicating similar perturbation difficulty. For instance, GPT-4 achieves ROP of 0.879 on \textit{KnInvPara} and 0.909 on \textit{RewriteQ}. Similarly, different tolerance levels are observed for surface noise: GPT-4 scores 0.960 under \textit{OptionForm} but drops to 0.733 under \textit{InsertIrrChars}, showing that even subtle variations can significantly impact performance.

Our additional transformations further stress the models and reveal robustness gaps. The ``Others'' category introduces more complex perturbations, leading to lower ROPs for weaker models, such as Llama-3.1-8B with an average ROP of 0.418, while stronger models like GPT-4 and DeepSeek-R1 maintain high consistency (0.799–0.900). This demonstrates that our framework provides a more comprehensive and discriminative robustness evaluation than PertEval.

\subsection{RQ3: Impact of Combining Evolutionary Operations on Model Robustness}
\label{sec:rq3}

\begin{table}[htbp]
\caption{Effects of Using Two Evolutionary Methods on LLMs}\label{rq3_results}
\centering
\resizebox{\textwidth}{!}{ % 控制表格整体宽度
\begin{tabular}{@{} l *{8}{c} c @{}}  % 所有数字列改为居中对齐
\toprule
Method & \multicolumn{8}{c}{Model Performance} & AVG \\ 
\cmidrule(lr){2-9}  % 中间分隔线
       & DeepSeek-R1 & DeepSeek-V3 & Gemini-1.5 & GLM-4 & Llama-3.1 & Mistral-small & GPT-3.5 & GPT-4 & \\ 
\midrule
Origin         &  85.232 &  89.451 &  81.857 &  82.700 &  29.451 &  75.105 &  82.700 &  83.122 & -- \\
\midrule
AddIrrOpts            &  -0.844 &  -2.110 &  -8.861 &  -2.954 & -10.928 &  -1.266 &  -2.532 &  -2.110 & -3.950 \\
AddStrongDist            &  -7.890 &  -7.173 &  -9.283 &  -5.781 &  -8.059 &  -9.705 &  -5.781 &  -6.203 & -7.484 \\
ShuffleOptIds            &  -1.688 &  -5.485 &  -8.861 &  -1.266 & -13.671 &   0.000 &  -1.266 &   0.844 & -3.924 \\
ShuffleOptOrder            &  -0.844 &  -3.797 & -10.549 &   0.422 & -12.194 &   3.376 &   0.844 &  -0.422 & -2.896 \\
AddIrrOpts+AddIrrOpts        &  -4.515 &  -5.907 & -10.844 &  -7.468 &  -8.819 &  -4.093 &  -6.624 &  -6.203 & -6.809 \\
AddIrrOpts+ShuffleOptIds        &   1.266 &  -4.641 & -24.473 &  -4.219 & -20.464 &  -8.017 &  -4.219 &  -4.219 & -8.623 \\
AddIrrOpts+ShuffleOptOrder        &  -3.376 &  -5.485 & -20.253 &  -6.751 & -23.038 &  -7.173 &  -4.641 &  -5.485 & -9.525 \\
AddStrongDist+AddIrrOpts        &  -4.641 &  -5.485 & -13.080 &  -7.595 &  -9.240 & -10.127 &  -9.283 &  -9.578 & -8.629 \\
AddStrongDist+AddStrongDist        &  -8.734 &  -4.641 & -13.797 &  -6.329 &  -5.865 & -11.814 &  -5.485 &  -6.329 & -7.874 \\
\midrule
AVG            &  -3.803 &  -5.327 & -13.892 &  -4.873 & -12.669 &  -5.944 &  -4.557 &  -4.699 & -- \\
\bottomrule
\end{tabular}
}

\end{table}

\parabf{Design.}
To explore how combinations of evolutionary operations affect LLM performance, we investigate whether applying two evolution operations in sequence leads to greater performance degradation compared to applying them individually. Building on the findings of RQ1, we selected several evolution operations that had demonstrated strong impact and constructed all possible pairwise combinations. Each input instance is subjected to two consecutive transformations (e.g., \textit{AddIrrOpts} followed by \textit{ShuffleOptOrder}). This setup simulates more realistic scenarios where adversarial inputs may undergo multiple, compounding modifications.

\parabf{Results and Analysis.}
As shown in Table~\ref{rq3_results}, even a single evolution operation causes noticeable performance drops, with average degradation ranging from $-2.896$ to $-7.484$. Among the individual operations, \textit{AddStrongDist} induces the largest performance decline, highlighting its effectiveness in confusing models with plausible distractors.

When two evolution operations are applied in combination, the performance degradation becomes significantly more severe. For instance, the sequence \textit{AddIrrOpts} $\rightarrow$ \textit{ShuffleOptOrder} results in an average accuracy drop of $-9.525$, while applying \textit{AddStrongDist} twice leads to a decrease of $-7.874$. Even applying the same operation twice (e.g., \textit{AddIrrOpts} $\rightarrow$ \textit{AddIrrOpts}) causes a substantial decline of $-6.809$. \textbf{These results suggest that combining operations—whether different or repeated—has a compounding negative effect on model performance.} Importantly, the total degradation from two operations is often greater than the sum of their individual effects, indicating a nonlinear compounding effect. This is especially evident when the operations impact both the semantic content and structural layout of the input. For example, injecting irrelevant options and then shuffling their order disrupts both the informational clarity and the visual structure that the model may rely on.

We also observe that models with weaker baseline robustness are more vulnerable to combined operations. \textbf{LLaMA-3.1-8B}, which has the lowest original accuracy (29.451), experiences the most severe performance drop ($-12.669$). In contrast, stronger models such as \textbf{GPT-4} and \textbf{DeepSeek-R1} exhibit more limited declines ($-4.699$ and $-3.803$, respectively), although they are not immune to degradation.

\textbf{In summary, RQ3 confirms that combining two evolutionary operations significantly amplifies the adversarial effect, especially when those operations target diverse aspects of the input.} This validates the utility of multi-step evolution strategies in exposing vulnerabilities that may be missed under single-step perturbations.
\subsection{RQ4: Impact of Long Evolution Chains on Performance}
\label{sec:rq4}

\parabf{Design}
To simulate more realistic and adversarial input conditions, we constructed long evolution chains by sequentially applying multiple strong atomic operations identified in RQ1. We designed three fixed pipelines: a Rule-based chain applying five structural operations (\textit{UpdateOptIds}, \textit{ShuffleOptOrder}, \textit{InsertIrrChars}, \textit{AddAboveWrong}, and \textit{SwapQOpt}); an LLM-based chain using five semantic operations via LLMs (\textit{RewriteOptRAG}, \textit{AddStrongDist}, \textit{RewriteQ}, \textit{AbbrQ}, and \textit{TransQEnZh}); and a Hybrid chain combining both types, including \textit{RewriteOpt} (twice), \textit{AddIrrOpts}, \textit{AddStrongDist}, and \textit{ShuffleOptOrder}. Each chain was applied in a fixed order to all dataset instances, resulting in three deeply evolved versions. This setup aims to examine how increasing transformation depth affects model robustness.

\parabf{Results and Analysis}
As shown in Table~\ref{rq4_results}, all long evolution chains caused significant performance drops, confirming that cumulative perturbations severely undermine LLM robustness.

The Rule-based pipeline led to the largest degradation, with an average drop of $-52.93$. Several models—\texttt{GGGemini-1.5}, \texttt{Mistral-small}, and both GPT variants—saw declines over 68 points, confirming that accumulated structural noise can severely impair model comprehension. For example, \texttt{DeepSeek-V3} and \texttt{GLM-4} dropped by $-45.15$ and $-68.35$, respectively.

The LLM-based pipeline caused a moderate but still notable average drop of $-28.76$. While semantic-level operations like paraphrasing or translation are less disruptive individually, their repeated application still significantly affects accuracy, revealing the limits of robustness under semantic perturbation.

The Hybrid pipeline showed the smallest average drop ($-22.97$), but the degradation remains considerable. As the number of evolution rounds increases and more diverse operations are combined, further performance declines are likely—highlighting the potential of complex, chained perturbations for revealing deeper weaknesses in LLM robustness.

Robustness also varied significantly across models. \texttt{Llama-3.1-8B}, which had the lowest baseline accuracy, exhibited smaller declines under LLM-based and hybrid chains ($-7.26$ and $-17.64$), likely due to limited performance ceiling. In contrast, stronger models like \texttt{Gemini-1.5} and \texttt{GPT-4} were more affected by the Rule pipeline, highlighting that even top-tier systems are susceptible to deep structural corruption.

\textbf{Overall, this experiment confirms that both the length and type of evolution chains significantly impact model robustness.} Long evolution pipelines serve as effective stress tests, uncovering hidden vulnerabilities not visible under single-step evaluation.

\begin{table}[htbp]
\caption{Long Evolution Chains on Performance. Rule = UpdateOptIds + ShuffleOptOrder + InsertIrrChars + AddAboveWrong + SwapQOpt; 
LLM = RewriteOptRAG + AddStrongDist + RevQ + AbbrQ + TransQEnZh; 
Rule+LLM =RewriteOpt + AddIrrOpts + AddStrongDist + ShuffleOptOrder.}
\label{rq4_results}
\centering
\resizebox{\textwidth}{!}{ % 控制表格整体宽度
\begin{tabular}{@{} l *{8}{c} c @{}}  % 所有数值列居中对齐
\toprule
Method & \multicolumn{8}{c}{Model Performance} & AVG \\ 
\cmidrule(lr){2-9}
       & DeepSeek-R1 & DeepSeek-V3 & Gemini-1.5 & GLM-4 & Llama-3.1 & Mistral-small & GPT-3.5 & GPT-4 & \\ 
\midrule
Origin    &  85.232 &  89.451 &  81.857 &  82.700 &  29.451 &  75.105 &  82.700 &  83.122 &  --    \\
Rule      &  -6.751 & -45.148 & -71.308 & -68.354 & -23.882 & -70.886 & -68.776 & -68.354 & -52.932 \\
LLM       & -27.848 & -27.848 & -43.038 & -29.114 &  -7.257 & -37.553 & -28.692 & -28.692 & -28.755 \\
Rule+LLM  & -18.017 & -24.219 & -35.865 & -21.688 & -17.637 & -23.376 & -20.844 & -22.110 & -22.969 \\
\bottomrule
\end{tabular}
}

\end{table}

\section{DISCUSSION}
\label{sec:discussion}

\parabf{Conclusion}
We propose \app, an evolution-based evaluation framework for close-ended tasks, featuring 22 atomic evolution operations and multi-round composition to simulate diverse perturbations. Our experiments show that operations disrupting logical structure or adding misleading semantics most effectively test robustness, causing an average accuracy drop of 7.283. Models vary in sensitivity, and \app provides more comprehensive robustness evaluation than prior methods. Combining evolution operations amplifies adversarial effects, with both operation type and chain length significantly impacting performance. These results highlight the need for evolution-aware robustness evaluation in large language models.

\parabf{Limitations and Future Work}
While our findings are reproducible and interpretable, several limitations remain. First, the selected models—though diverse—do not cover all architectures or training paradigms, limiting generalizability. Second, our benchmark focuses on multi-choice QA; broader task coverage (e.g., reasoning, summarization) is needed. Third, the evolution operations, while representative, are not exhaustive—other natural or adversarial changes (e.g., syntactic rephrasing, factual errors) remain unexplored. Moreover, LLM-driven evolutions involve prompt design and randomness, introducing variability despite our efforts to control it. Lastly, our evaluation emphasizes answer accuracy, which may overlook nuanced behavioral shifts. Future work will expand our framework to more tasks and evolution types, explore automated evolution-repair cycles, and integrate richer evaluation metrics (e.g., confidence, fairness), with the ultimate goal of advancing the development of trustworthy and evolution-resilient LLMs.

\bibliographystyle{plainnat}  % 文献样式，可以改为 ieeetr、alpha 等
\bibliography{ref}  % references.bib 是你的 bib 文件名（不要写后缀）
\newpage
\appendix
\section{Methods}

\textbf{Origin} Question: According to Piaget, children are \_\_\_\_. Options: (A) ``Blank slates'', (B) Less intelligent than adults, (C) ``Little scientists'', (D) Shaped by culture \\

\textbf{AbbreviateOptionsContent} Question: According to Piaget, children are \_\_\_\_. Options: (A) Blank slates, (B) Less intelligent than adults, (C) Little scientists, (D) Shaped by culture \\

\textbf{AbbreviateQuestion} According to Piaget, children are? Options: (A) ``Blank slates'', (B) Less intelligent than adults, (C) ``Little scientists'', (D) Shaped by culture \\

\textbf{AddAboveWrongOptions} Question: According to Piaget, children are \_\_\_\_. Options: (A) ``Blank slates'', (B) Less intelligent than adults, (D) Shaped by culture, (C) None of the above \\

\textbf{AddIrrelatedOptions} Question: According to Piaget, children are \_\_\_\_. Options: (A) ``Blank slates'', (B) Less intelligent than adults, (C) ``Little scientists'', (D) Shaped by culture, (E) Experts in quantum physics, (F) More interested in Li Bai's poems than empirical observation \\

\textbf{AddStrongDistractors} Question: According to Piaget, children are \_\_\_\_. Options: (A) ``Blank slates'', (B) Less intelligent than adults, (C) ``Little scientists'', (D) Shaped by culture, (E) Always bound to utilitarian decision-making, (F) Influenced by astrophysical constants, (G) Not unrelated to the concept of cognitive dissonance \\

\textbf{ConvertCorrectOptionToJudge} Question: According to Piaget, children are ``little scientists''. \\

\textbf{ConvertWrongOptionsToJudge} Question: According to Piaget, children are ``blank slates''. \\

\textbf{ExpandOptionsWithIrrelatedInfo} Question: According to Piaget, children are \_\_\_\_. Options: (A) ``Blank slates'' (some theories suggest this), (B) Less intelligent than adults (a common misconception), (C) ``Little scientists'' (supported by Piaget’s research in 1952), (D) Shaped by culture (though not Piaget’s primary focus) \\

\textbf{ExpandOptionsWithRelatedInfo} Question: According to Piaget, children are \_\_\_\_. Options: (A) ``Blank slates'' (implies children are born without innate knowledge or abilities, which contradicts Piaget’s theory of cognitive development), (B) Less intelligent than adults (misrepresents Piaget’s view; he emphasized qualitative differences in thinking, not inferior intelligence), (C) ``Little scientists'' (reflects Piaget’s belief that children actively construct knowledge through exploration and experimentation), (D) Shaped by culture (while culture influences development, this does not align with Piaget’s focus on universal stages of cognitive growth) \\

\textbf{ExpandQuestionWithIrrelatedInfo} Question: According to Piaget, as cited in the \textit{Journal of Developmental Psychology} (2023), children exhibit a unique cognitive framework that aligns with which of the following descriptions? Additionally, under standard developmental conditions (pH=7.2, 25℃), WHO experts emphasize the importance of understanding this framework. Options: (A) ``Blank slates'', (B) Less intelligent than adults, (C) ``Little scientists'', (D) Shaped by culture \\

\textbf{ExpandQuestionWithRelatedInfo} Question: In the context of Jean Piaget's developmental psychology, which term best describes how he characterized children's cognitive processes and their approach to understanding the world? Options: (A) ``Blank slates'', (B) Less intelligent than adults, (C) ``Little scientists'', (D) Shaped by culture \\

\textbf{InsertIrrelevantCharacters} Question: Ac\#c\#o\%rding \&\#to Pia\#get, children are \_\_@\_\_\_\_\%\&\_. Options: (A) ``Blank slates'', (B) Less intelligent than adults, (C) ``Little scientists'', (D) Shaped by culture \\

\textbf{ReverseQuestion} Question: According to Piaget, children are not \_\_\_\_. Options: (A) ``Blank slates'', (B) Less intelligent than adults, (C) ``Little scientists'', (D) Shaped by culture \\

\textbf{RewriteOptionsContent} Question: According to Piaget, children are \_\_\_\_. Options: (A) ``Empty canvases'', (B) Not as smart as grown-ups, (C) ``Mini researchers'', (D) Influenced by societal norms \\

\textbf{RewriteOptionsUsingRAG} Question: According to Piaget, children are \_\_\_\_. Options: (A) ``Blank slates'', (B) Less intelligent than adults, (C) ``Little scientists'', (D) Shaped by culture \\

\textbf{RewriteQuestion} Question: According to Piaget's theory, how are children best described? Options: (A) ``Blank slates'', (B) Less intelligent than adults, (C) ``Little scientists'', (D) Shaped by culture \\

\textbf{RewriteQuestionUsingRAG} Question: According to Piaget, children’s cognitive development progresses through stages, with the sensorimotor stage (birth to 2 years) marked by the emergence of object permanence and the preoperational stage (2 to 7 years) characterized by limitations such as centration, irreversibility, and egocentrism. Options: (A) ``Blank slates'', (B) Less intelligent than adults, (C) ``Little scientists'', (D) Shaped by culture \\

\textbf{ShuffleOptionIds} Question: According to Piaget, children are \_\_\_\_. Options: (B) ``Blank slates'', (D) Less intelligent than adults, (C) ``Little scientists'', (A) Shaped by culture \\

\textbf{ShuffleOptionsOrder} Question: According to Piaget, children are \_\_\_\_. Options: (C) ``Little scientists'', (D) Shaped by culture, (A) ``Blank slates'', (B) Less intelligent than adults \\

\textbf{SwapQuestionWithOptions} Options: (A) ``Blank slates'', (B) Less intelligent than adults, (C) ``Little scientists'', (D) Shaped by culture. Question: According to Piaget, children are \_\_\_\_. \\

\begin{CJK}{UTF8}{gbsn}
\textbf{TranslateOptionsEnZh} Question: According to Piaget, children are \_\_\_\_. Options: (A) ``白板'', (B) 不如成年人聪明, (C) ``小科学家'', (D) 受文化影响 \\
\textbf{TranslateQuestionEnZh} Question: 根据皮亚杰的观点，儿童是\_\_\_\_。Options: (A) ``Blank slates'', (B) Less intelligent than adults, (C) ``Little scientists'', (D) Shaped by culture \\
\end{CJK}

\section{Prompt Design for Model-Based Evolution}
\label{appendix:atomic_operations}
\subsection{Prompt template for evolution operations}
\begin{figure}[htpb]
    \centering
    \includegraphics[width=0.7\linewidth]{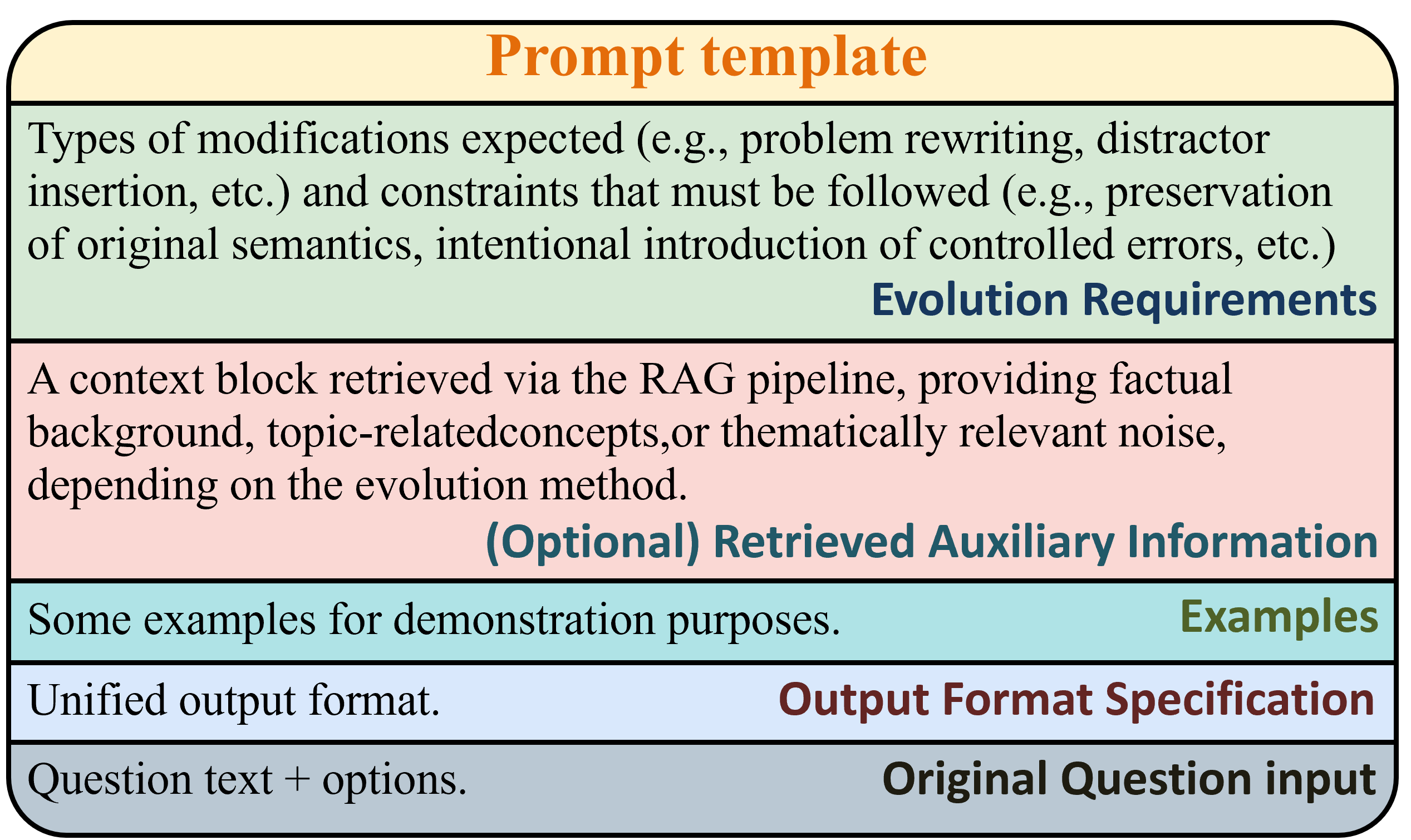}
    \caption{Prompt template, all Type LLM evolution method implementation can refer to this template}
    \label{fig:prompt_template}
\end{figure}

To effectively implement evolutionary data augmentation using large language models (LLMs), we abstract the prompt structure into a modular template that guides the model through each transformation process. Specifically, each evolution prompt is composed of four key components:

\textbf{Evolution Requirements}
This section outlines the core principles and key considerations for the given evolution method. It clearly defines what type of modification is expected (e.g., question rewriting, distractor insertion) and what constraints must be adhered to (e.g., preserving original semantics, maintaining answer correctness or intentionally introducing controlled errors).

\textbf{(Optional) Retrieved Auxiliary Information}
A context block retrieved via the RAG pipeline, providing factual background, topic-related concepts, or thematically relevant noise, depending on the evolution method.

\textbf{Examples}
We provide several illustrative examples demonstrating how the evolution should be performed. These examples serve to ground the model's understanding and help it generalize the transformation pattern appropriately to new inputs.

\textbf{Output Format Specification}
To ensure consistent and structured responses from the LLM, we explicitly define the required output format. 

\textbf{Original Question Input}
The final part of the prompt is the original input to be evolved. This includes:

Question Text: The textual content of the original question.

Options: A list of candidate options, typically labeled (A), (B), (C), etc., with one or more correct answers.

By structuring the prompts in this way, we ensure that the large model can perform controlled and consistent evolutions, enabling us to generate high-quality, diverse, and targeted test instances for LLM evaluation.

\section{Models}
\label{appendix:models}
\begin{table}[ht]
\centering
\caption{List of evaluated models with open-source status, scale, release date, and company}
\begin{tabular}{|c|c|c|c|c|}
\hline
\textbf{Model Name} & \textbf{Is Open Source} & \textbf{Scale} & \textbf{Release Date} & \textbf{Company} \\
\hline
GPT-4 & No & - & 2023 & OpenAI \\
GPT-3.5 & No & - & 2022 & OpenAI \\
Gemini 1.5 & No & - & 2024 & Google DeepMind \\
GLM-4 & Yes & - & 2024 & Tsinghua University \\
Mistral-small & Yes & 24B & 2023 & Mistral AI \\
LLaMA-3.1 & Yes & 8B & 2023 & Meta \\
DeepSeek-R1 & No & 671B & 2023 & DeepSeek AI \\
DeepSeek-V3 & No & 671B & 2024 & DeepSeek AI \\
\hline
\end{tabular}
\end{table}

\section{RQ1: Dataset result detail}
\label{rq1_dataset__detail_results}
\begin{table}[htbp]
\caption{History Dataset Results}\label{tab:rq1_history}
\centering
\resizebox{\textwidth}{!}{ % 保持原始数据精度
\begin{tabular}{@{} l *{8}{c} c @{}}  % 严格保持原始数值
\toprule
\multicolumn{1}{c}{Method} & \multicolumn{8}{c}{Model Performance} & AVG \\ 
\cmidrule(lr){2-9}
       & DeepSeek-R1 & DeepSeek-V3 & Gemini-1.5 & GLM-4 & Llama-3.1 & Mistral-small & GPT-3.5 & GPT-4 & \\ 
\midrule
Origin                              & 85.23  & 89.45  & 81.86  & 82.70  & 29.45  & 75.11  & 82.70  & 83.12  & -- \\
AbbrOptCont            & -6.33  & 0.42   & -3.38  & 0.42   & -2.49  & -3.38  & 0.42   & 0.00   & -1.788 \\
\textbf{AbbrQ}         & -2.11  & -6.33  & -9.28  & -8.02  & 2.19   & -6.75  & -5.49  & -6.33  & \textbf{-5.264} \\
\textbf{AddAboveWrong}       & -5.06  & -7.17  & -23.63 & -33.33 & -4.14  & -14.77 & -29.54 & -35.44 & \textbf{-19.135} \\
AddIrrOpts                 & -0.84  & -2.11  & -8.86  & -2.95  & -10.93 & -1.27  & -2.53  & -2.11  & -3.950 \\
\textbf{AddStrongDist}       & -7.89  & -7.17  & -9.28  & -5.78  & -8.06  & -9.70  & -5.78  & -6.20  & \textbf{-7.484} \\
\textbf{OptToJudge}       & -13.50 & -21.10 & -17.31 & -22.79 & -4.14  & -21.53 & -19.41 & -21.52 & \textbf{-17.663} \\
ExpandOptsIrr      & 1.69   & 0.42   & 4.64   & 4.64   & 8.86   & 5.91   & 4.22   & 2.95   & 4.167 \\
ExpandOptsRel        & -2.53  & 3.38   & 8.44   & 5.91   & 2.95   & 9.70   & 6.75   & 5.91   & 5.063 \\
ExpandQuesIrr     & 0.42   & 0.00   & 0.42   & 1.27   & 2.87   & -2.53  & 0.84   & 1.27   & 0.570 \\
ExpandQuesRel       & 0.42   & 1.27   & 4.64   & 5.06   & 2.62   & 3.80   & 3.80   & 4.22   & 3.228 \\
\textbf{InsertIrrChars} & -9.70  & -4.22  & -1.69  & -7.17  & -5.02  & -10.55 & -6.75  & -5.91  & \textbf{-6.377} \\
\textbf{RevQ}            & -59.07 & -53.92 & -54.60 & -53.88 & -17.43 & -53.16 & -53.21 & -53.80 & \textbf{-49.884} \\
RewriteOpti               & -2.53  & -2.95  & -2.11  & -4.22  & -2.91  & -4.22  & -1.69  & -4.22  & -3.107 \\
\textbf{RewriteOptRAG}     & -11.39 & -11.81 & -8.44  & -13.92 & -4.64  & -18.14 & -16.03 & -15.19 & \textbf{-12.447} \\
RewriteQ                     & -2.11  & -5.06  & -2.11  & -0.42  & -1.94  & -2.53  & 0.84   & -0.42  & -1.719 \\
RewriteQRAG             & -2.53  & 0.00   & 0.84   & -0.84  & 1.01   & -1.69  & -1.27  & -1.27  & -0.717 \\
ShuffleOptIds                    & -1.69  & -5.49  & -8.86  & -1.27  & -13.67 & 0.00   & -1.27  & 0.84   & -3.924 \\
ShuffleOptOrder                 & -0.84  & -3.80  & -10.55 & 0.42   & -12.19 & 3.38   & 0.84   & -0.42  & -2.896 \\
\textbf{SwapQOpt}    & -6.33  & -5.91  & -15.19 & -10.97 & -16.24 & -20.68 & -10.97 & -12.66 & \textbf{-12.368} \\
TransOptEnZh                & -0.84  & -2.11  & -0.84  & -2.53  & 11.52  & 0.00   & -2.53  & -2.53  & 0.016 \\
TransQEnZh               & 0.42   & -5.49  & -21.94 & -3.38  & 8.19   & -5.49  & -3.38  & -2.53  & -4.198 \\
UpdateOptIds                     & 2.95   & -2.95  & -5.91  & -2.95  & -22.28 & -1.27  & -1.69  & -4.22  & -4.789 \\
\midrule
AVG                                 & -5.410 & -6.431 & -7.708 & -6.846 & -2.059 & -6.892 & -6.150 & -6.878 & -6.047 \\
\bottomrule
\end{tabular}
}
\end{table}

\begin{table}[htbp]
\caption{Math Dataset Results}\label{tab:rq1_math}
\centering
\resizebox{\textwidth}{!}{
\begin{tabular}{@{} l *{8}{c} c @{}}
\toprule
\multicolumn{1}{c}{Method} & \multicolumn{8}{c}{Model Performance} & AVG \\ 
\cmidrule(lr){2-9}
       & DeepSeek-R1 & DeepSeek-V3 & Gemini-1.5 & GLM-4 & Llama-3.1 & Mistral-small & GPT-3.5 & GPT-4 & \\ 
\midrule
Origin                              & 95.00  & 63.00  & 66.00  & 44.00  & 14.00  & 24.00  & 43.00  & 43.00  & -- \\
AbbrOptCont            & -7.00  & -2.00  & 3.00   & 0.00   & -1.00  & 2.00   & 1.00   & 3.00   & -0.125 \\
AbbrQ         & -15.00 & -5.00  & -8.00  & 1.00   & -3.00  & 4.00   & -1.00  & 3.00   & -3.000 \\
\textbf{AddAboveWrong}       & -10.00 & -27.00 & -54.00 & -29.00 & -10.00 & -17.00 & -26.00 & -27.00 & \textbf{-25.000} \\
AddIrrOpts                 & -7.00  & -3.00  & -9.00  & -3.00  & -5.00  & -4.00  & -1.00  & -6.00  & -4.750 \\
AddStrongDist                & -3.00  & -5.00  & -6.00  & -9.00  & 0.00   & -3.00  & -5.00  & -8.00  & -4.875 \\
\textbf{OptToJudge}       & -8.00  & -35.00 & -23.00 & -24.00 & 5.00   & 7.00   & -21.00 & -19.00 & \textbf{-14.750} \\
ExpandOptsIrr      & -5.00  & 12.00  & 2.00   & 23.00  & 32.00  & 37.00  & 28.00  & 26.00  & 19.375 \\
ExpandOptsRel        & -8.00  & 11.00  & 10.00  & 25.00  & 21.00  & 31.00  & 21.00  & 26.00  & 17.125 \\
ExpandQuesIrr     & -11.00 & -6.00  & -1.00  & -1.00  & -6.00  & -4.00  & 1.00   & 1.00   & -3.375 \\
ExpandQuesRel       & -17.00 & -1.00  & -4.00  & -2.00  & 1.00   & -4.00  & 4.00   & -5.00  & -3.500 \\
\textbf{InsertIrrChars} & -9.00  & -16.00 & -12.40 & -13.00 & -10.00 & -5.00  & -13.00 & -8.00  & \textbf{-10.800} \\
\textbf{RevQ}            & -58.50 & -37.80 & -44.90 & -20.60 & -3.50  & -6.60  & -19.30 & -19.80 & \textbf{-26.375} \\
RewriteOpti               & -8.00  & -5.00  & 2.00   & 7.00   & -3.00  & 2.00   & 2.00   & 0.00   & -0.375 \\
RewriteQ                     & -11.00 & -5.00  & -4.00  & 3.00   & 1.00   & 0.00   & 5.00   & 3.00   & -1.000 \\
ShuffleOptIds                    & -4.00  & -9.00  & -2.00  & -6.00  & -4.00  & 3.00   & 0.00   & 1.00   & -2.625 \\
ShuffleOptOrder                 & -9.00  & -12.00 & -11.00 & -7.00  & -7.00  & 5.00   & -1.00  & -2.00  & -5.500 \\
SwapQOpt    & -7.00  & -31.00 & -34.00 & -6.00  & -6.00  & -5.00  & -9.00  & -6.00  & -13.000 \\
TransOptEnZh                & -8.00  & -16.00 & -9.00  & 0.00   & 4.00   & -1.00  & 4.00   & 1.00   & -3.125 \\
TransQEnZh               & -34.00 & -5.00  & -8.00  & -6.00  & 3.00   & -3.00  & -3.00  & -2.00  & -7.250 \\
UpdateOptIds                     & -16.00 & -9.00  & -4.00  & -4.00  & 0.00   & 3.00   & 3.00   & -2.00  & -3.625 \\
\midrule
AVG                                 & -12.775 & -10.340 & -10.865 & -3.580 & 0.425 & 2.070 & -1.515 & -2.040 & -4.828 \\
\bottomrule
\end{tabular}
}
\end{table}

\begin{table}[htbp]
\caption{Medicine Dataset Results}\label{tab:rq1_medicine}
\centering
\resizebox{\textwidth}{!}{
\begin{tabular}{@{} l *{8}{c} c @{}}
\toprule
\multicolumn{1}{c}{Method} & \multicolumn{8}{c}{Model Performance} & AVG \\ 
\cmidrule(lr){2-9}
       & DeepSeek-R1 & DeepSeek-V3 & Gemini-1.5 & GLM-4 & Llama-3.1 & Mistral-small & GPT-3.5 & GPT-4 & \\ 
\midrule
Origin                              & 95.88  & 87.86  & 79.77  & 86.76  & 18.34  & 58.28  & 85.29  & 86.76  & -- \\
AbbrOptCont            & -5.44  & -0.36  & -0.36  & -0.37  & -1.14  & -0.93  & 0.37   & 0.37   & -0.983 \\
\textbf{AbbrQ}         & -10.59 & -8.45  & -15.07 & -9.56  & -4.78  & -4.24  & -8.46  & -8.82  & \textbf{-8.746} \\
\textbf{AddAboveWrong}       & -21.62 & -27.94 & -61.39 & -60.29 & -13.19 & -31.08 & -57.72 & -62.50 & \textbf{-41.966} \\
AddIrrOpts                 & -5.08  & 1.47   & -8.82  & -0.74  & -8.97  & 5.32   & 0.00   & -0.74  & -2.195 \\
AddStrongDist                & -5.08  & 1.11   & -13.26 & -1.10  & -5.04  & 2.01   & 0.37   & 0.00   & -2.624 \\
\textbf{OptToJudge}       & -12.06 & -13.60 & -5.88  & -27.94 & 27.61  & 3.48   & -21.69 & -25.37 & \textbf{-9.431} \\
ExpandOptsIrr      & -6.18  & 7.72   & 10.67  & 6.25   & 7.72   & 30.69  & 7.35   & 4.51   & 8.591 \\
ExpandOptsRel        & -5.08  & 6.99   & 11.40  & 6.25   & 10.26  & 27.38  & 8.09   & 5.88   & 8.896 \\
ExpandQuesIrr     & -5.08  & 0.37   & -3.67  & -0.37  & 0.22   & -0.93  & 0.37   & -2.18  & -1.409 \\
ExpandQuesRel       & -6.91  & 1.47   & 6.99   & 0.74   & 6.54   & 7.52   & 1.84   & 1.47   & 2.458 \\
\textbf{InsertIrrChars} & -4.71  & -9.92  & -8.45  & -22.06 & -4.71  & -20.42 & -21.69 & -21.69 & \textbf{-14.206} \\
\textbf{RevQ}            & -66.29 & -56.50 & -51.61 & -58.31 & -9.30  & -38.73 & -57.17 & -58.64 & \textbf{-49.569} \\
RewriteOpti               & -8.75  & -4.78  & -5.51  & -6.99  & -6.87  & -6.08  & -4.41  & -5.88  & -6.159 \\
\textbf{RewriteOptRAG}     & -20.52 & -20.59 & -41.91 & -22.79 & -0.29  & -11.59 & -23.16 & -23.53 & \textbf{-20.548} \\
RewriteQ                     & -7.63  & -0.36  & 2.21   & -2.94  & -0.07  & 0.54   & -1.84  & -4.41  & -1.813 \\
RewriteQRAG             & -5.44  & 0.37   & -20.22 & -2.94  & -3.05  & 3.48   & -1.84  & -2.94  & -4.073 \\
ShuffleOptIds                    & -9.49  & 0.00   & -9.55  & 1.84   & -10.88 & 9.36   & 0.73   & -0.37  & -2.295 \\
ShuffleOptOrder                 & -7.28  & -3.67  & -11.39 & -0.37  & -11.03 & 10.10  & 3.31   & 0.37   & -2.495 \\
\textbf{SwapQOpt}    & -8.38  & -3.67  & -4.77  & -17.28 & -11.73 & -10.49 & -16.18 & -17.65 & \textbf{-11.269} \\
TransOptEnZh                & -5.08  & 0.74   & -2.94  & -6.99  & 7.43   & -4.24  & -4.41  & -6.25  & -2.718 \\
\textbf{TransQEnZh}      & -8.38  & -1.47  & -46.32 & -6.62  & -2.54  & -5.34  & -6.25  & -7.35  & \textbf{-10.534} \\
UpdateOptIds                     & -3.61  & -1.47  & -8.45  & -3.31  & -15.04 & 4.22   & -1.47  & -3.31  & -4.055 \\
\midrule
AVG                                 & -10.307 & -5.675 & -11.430 & -10.472 & 2.104 & 0.466 & -8.847 & -10.526 & -6.836 \\
\bottomrule
\end{tabular}
}
\end{table}

\begin{table}[htbp]
\caption{Psychology Dataset Results}\label{tab:rq1_psychology}
\centering
\resizebox{\textwidth}{!}{
\begin{tabular}{@{} l *{8}{c} c @{}}
\toprule
\multicolumn{1}{c}{Method} & \multicolumn{8}{c}{Model Performance} & AVG \\ 
\cmidrule(lr){2-9}
       & DeepSeek-R1 & DeepSeek-V3 & Gemini-1.5 & GLM-4 & Llama-3.1 & Mistral-small & GPT-3.5 & GPT-4 & \\ 
\midrule
Origin                              & 89.54  & 83.49  & 73.69  & 78.92  & 29.08  & 67.32  & 77.77  & 78.59  & -- \\
AbbrOptCont            & -1.64  & -3.43  & -4.41  & -2.29  & -4.58  & -2.62  & -2.61  & -3.10  & -3.085 \\
\textbf{AbbrQ}         & -3.60  & -7.02  & -36.44 & -7.19  & -1.80  & -6.38  & -6.37  & -5.23  & \textbf{-9.254} \\
\textbf{AddAboveWrong}       & -6.16  & -24.18 & -40.85 & -56.21 & -19.68 & -37.59 & -55.06 & -55.39 & \textbf{-36.890} \\
AddIrrOpts                 & -2.29  & -3.92  & -10.13 & -3.76  & -1.21  & -6.05  & -2.45  & -3.59  & -4.175 \\
AddStrongDist                & -3.11  & -4.57  & -9.81  & -3.92  & 2.52   & -5.23  & -2.94  & -4.09  & -3.894 \\
\textbf{OptToJudge}       & -18.04 & -20.26 & -10.79 & -21.12 & 9.97   & -18.47 & -20.75 & -19.44 & \textbf{-14.863} \\
ExpandOptsIrr      & -0.33  & 2.62   & 4.74   & 5.06   & 8.99   & 9.15   & 7.52   & 6.54   & 5.536 \\
ExpandOptsRel        & -1.96  & 5.72   & 9.47   & 7.68   & 4.25   & 9.47   & 8.50   & 7.19   & 6.290 \\
ExpandQuesIrr     & -7.52  & -2.29  & -2.45  & -2.94  & 2.29   & -4.09  & -1.63  & -1.31  & -2.493 \\
ExpandQuesRel       & -2.29  & 1.31   & 5.72   & 4.41   & 10.46  & 4.90   & 5.23   & 5.08   & 4.352 \\
\textbf{InsertIrrChars} & -0.98  & -12.74 & -10.95 & -17.32 & -10.29 & -22.06 & -14.54 & -16.34 & \textbf{-13.153} \\
\textbf{RevQ}            & -62.85 & -58.09 & -50.21 & -52.90 & -18.95 & -45.64 & -52.02 & -53.32 & \textbf{-49.248} \\
RewriteOpti               & -4.09  & -6.21  & -36.60 & -3.60  & -4.08  & -5.07  & -2.94  & -3.27  & -8.233 \\
\textbf{RewriteOptRAG}     & -22.71 & -25.00 & -19.12 & -22.06 & -12.58 & -23.70 & -19.93 & -20.92 & \textbf{-20.753} \\
RewriteQ                     & -1.96  & -0.49  & -1.96  & -2.13  & 1.96   & 0.00   & -0.98  & -1.31  & -0.859 \\
RewriteQRAG             & 0.49   & -0.33  & 0.32   & -0.82  & 2.45   & -0.70  & 1.15   & 0.33   & 0.361 \\
ShuffleOptIds                    & 0.65   & -5.06  & -8.34  & -3.43  & -13.24 & -1.31  & -2.45  & -1.96  & -4.393 \\
ShuffleOptOrder                 & -0.17  & -4.41  & -7.36  & -2.29  & -15.69 & -4.86  & -0.65  & -1.96  & -4.674 \\
\textbf{SwapQOpt}    & -4.41  & -10.78 & -34.15 & -15.85 & -14.71 & -23.81 & -15.68 & -15.52 & \textbf{-16.864} \\
\textbf{TransOptEnZh}       & -4.25  & -7.35  & -3.27  & -9.48  & -0.16  & -8.01  & -7.41  & -9.48  & \textbf{-6.177} \\
\textbf{TransQEnZh}      & -5.88  & -2.94  & -31.70 & -8.83  & -6.70  & -10.95 & -8.66  & -9.15  & \textbf{-10.602} \\
UpdateOptIds                     & -0.82  & -4.57  & -36.44 & -4.42  & -22.55 & -4.74  & -2.94  & -4.20  & \textbf{-10.085} \\
\midrule
AVG                                 & -6.831 & -8.464 & -13.450 & -9.493 & -1.208 & -8.296 & -8.390 & -8.884 & -8.127 \\
\bottomrule
\end{tabular}
}
\end{table}

\newpage
\end{document}